%% file: MAIN.tex
\documentclass{article}
\usepackage[final]{neurips_2025}

\usepackage[utf8]{inputenc} % allow utf-8 input
\usepackage[T1]{fontenc}    % use 8-bit T1 fonts
\usepackage{hyperref}       % hyperlinks
\usepackage{url}            % simple URL typesetting
\usepackage{booktabs}       % professional-quality tables
\usepackage{amsfonts}       % blackboard math symbols
\usepackage{nicefrac}       % compact symbols for 1/2, etc.
\usepackage{microtype}      % microtypography
\usepackage{xcolor}         % colors
\usepackage{todonotes}
\usepackage{wrapfig}
\usepackage{fontawesome}

\input{packages}
\newcommand{\model}{\texttt{EigenShift}}

\title{Redefining Experts: Interpretable Decomposition of Language Models for Toxicity Mitigation}

\author{Zuhair Hasan Shaik\thanks{Work done at FLaME-NLP Lab, IIIT Delhi.}\\IIIT Dharwad, India\\\texttt{zuhashaik12@gmail.com}\\
\And
Abdullah Mazhar\\IIIT Delhi, India\\\texttt{abdullahm@iiitd.ac.in}\\
\AND
Aseem Srivastava\\IIIT Delhi, India\\\texttt{aseems@iiitd.ac.in}\\
\And
Md. Shad Akhtar\\IIIT Delhi, India\\\texttt{shad.akhtar@iiitd.ac.in}}

\begin{document}

\maketitle

\begin{abstract}
Large Language Models have demonstrated impressive fluency across diverse tasks, yet their tendency to produce toxic content remains a critical challenge for AI safety and public trust. Existing toxicity mitigation approaches primarily manipulate individual neuron activations, but these methods suffer from instability, context dependence, and often compromise the model’s core language abilities. To address these shortcomings, we investigate three key questions: the stability of neuron-level toxicity indicators, the advantages of structural (layer-wise) representations, and the interpretability of mechanisms driving toxic generation. Through extensive experiments on \texttt{Jigsaw} and \texttt{ToxiCN} datasets, we show that aggregated layer-wise features provide more robust signals than single neurons. Moreover, we observe conceptual limitations in prior works that conflate toxicity \textit{detection} experts and \textit{generation} experts within neuron-based interventions. To mitigate this, we propose a novel principled intervention technique, \model, based on eigen-decomposition of the language model’s final output layer. This method selectively targets generation-aligned components, enabling precise toxicity suppression without impairing linguistic competence. Our method requires no additional training or fine-tuning, incurs minimal computational cost, and is grounded in rigorous theoretical analysis.

% Finally, we introduce the \textbf{TPH score}, a novel evaluation metric capturing the harmonic mean of toxicity reduction and perplexity preservation. Our approach achieves state-of-the-art performance on RealToxicPrompts and sets a new direction for safe, interpretable, and minimally invasive LLM editing.
\end{abstract}

\section{Introduction}
The rise of large language models has brought remarkable advancements across tasks involving human-like generation \cite{naveed2024comprehensiveoverviewlargelanguage, zhao2025surveylargelanguagemodels, minaee2025largelanguagemodelssurvey, brown2020languagemodelsfewshotlearners, srivastava-etal-2024-knowledge, 10.1145/3543507.3583380}. However, this progress has also led to a surge in toxic and unsafe machine-generated content, raising urgent concerns for AI safety and societal impact. Addressing this issue demands not only effective toxicity detection but also interventions that reduce toxic generation without hampering a model's fluency or general language understanding.

Prior work has primarily focused on neuron-level interventions, where specific neurons, often termed {\em concept experts}, are identified and manipulated based on their activation in response to toxic inputs \cite{whispering-experts, cuadros2022self, radford2017learning}. 
% While this line of work has yielded some success in toxicity reduction, it suffers from critical limitations. 
However, neuron activations are often stochastic and brittle, leading to inconsistent behavior across contexts \cite{whispering-experts}, which we also show in our analysis on the way. Furthermore, interventions at early layers propagate disruption downstream, resulting in catastrophic forgetting and significant degradation in language modeling performance \cite{10.5555/2987189.2987341}. To understand these aspects, we raise three research questions (RQs):
$\blacktriangleright$\textbf{RQ1:} Are individual neurons reliable indicators of toxicity, or do their activations merely reflect unstable correlations?
$\blacktriangleright$\textbf{RQ2:} Can layer-wise or structural representations capture toxicity more robustly, particularly across multilingual and specialized domains?
$\blacktriangleright$\textbf{RQ3:} Can we uncover interpretable components beyond just layers and neurons that contribute to toxic generation, aiming to make black box models more understandable?

% To address \textbf{RQ1} and \textbf{RQ2},
In our work, we conduct a comprehensive analysis using two toxicity detection datasets: \texttt{Jigsaw} (English) \cite{jigsaw2018toxic} and \texttt{ToxiCN} (Chinese) \cite{lu-etal-2023-facilitating}. We compare the reliability of individual neuron activations against aggregated layer-wise representations in detecting toxic content. 
% Our multilingual evaluation, based on AUROC, reveals that neuron-level activations are highly stochastic and often fail to capture toxicity consistently across languages, whereas layer-level signals exhibit more robust and semantically aligned behavior.
% For \textbf{RQ3}, 
Further, we delve deeper into the LLMs and identify the limited ability to distinguish between two crucial roles: (a) experts that detect toxicity, and (b) experts that generate toxic outputs. Most existing interventions conflate these two, suppressing both types of activations, which inadvertently damages the model’s ability to recognize toxicity, leading to catastrophic forgetting and degraded language performance \cite{10.5555/2987189.2987341}.

Hence, we propose the problem of controlling toxic generation while preserving the linguistic competence of LLMs. To address this, we propose a novel framework, \model. It targets the \textit{generation experts}, a set of interpretable eigen-components in the model's final decision space (LM head) while preserving the detection experts that are essential for identifying toxicity. We consider these principal vectors as the \textit{choices} of the LLM: in the absence of personality or intent, an LLM's behavior is governed by probability distributions. These eigen-components substantially influence the selection of output tokens given input prompts. By selectively controlling or steering these generation-specific directions, our eigen-decomposition-based strategy enables precise control over toxic outputs without disturbing the model's core language capabilities.
% Our approach provides a principled solution to toxicity reduction with minimal degradation in perplexity, overcoming long-standing limitations of neuron-centric methods and paving a new path for safe and interpretable LLM editing.
We evaluate our method on the standard \texttt{RealToxicPrompts} dataset and demonstrate that it generalizes better than prior approaches. Also, we notice that existing metrics fall short in quantifying the trade-off between toxicity mitigation and language modeling retention. Hence, we propose a new composite metric that takes the \textbf{H}armonic mean of \textbf{T}oxicity reduction and \textbf{P}erplexity preservation (TPH score). \model\ reduces toxicity by $58\%$ while minimal change in perplexity ($\Delta PPL=+3.62$), achieving the notable TPH Score of $60.37\%$. \model\ outperforms all baselines on TPH across all LLMs except MPT-7B, where its performance is on par with Aura. 
% Unlike prior methods that trade fluency for safety, \model strikes the best balance across all major LLMs.
Our contributions are summarized as follows\footnote{Code Repository: \faGithub\ \url{https://github.com/flamenlp/EigenShift}}:
\begin{itemize}[leftmargin=2em,noitemsep,nolistsep]
    \item We provide a comprehensive empirical analysis questioning the reliability of neurons in toxicity detection, demonstrating their stochastic and context-sensitive nature across languages.
    
    \item We propose a novel method, \model, a decomposition-based perspective that implements targeted intervention strategy, suppressing toxic outputs with \textit{generation experts} while preserving the detection capacity and overall linguistic competence.
    
    \item We propose a new metric, TPH score, a unified and interpretable evaluation metric based on the harmonic mean of toxicity reduction and perplexity preservation.
    
    \item We show state-of-the-art performance on the RealToxicPrompts benchmark, with substantial toxicity reduction and keeping low perplexity and setting a safer LLM intervention standard.
\end{itemize}

\section{Methodology}
\label{sec:method}
In this work, we address the problem of evaluating concept expertise within neural networks, specifically for detecting toxicity as a binary concept. The input to the network consists of text sentences $( S = (s_1, s_2, \dots, s_n) )$ from the dataset (S), where each ( $s_i$ ) represents a sentence in the input. The output is a binary classification vector $( y = (y_1, y_2, \dots, y_n) )$ where each ( $y_i$ ) is either 0 or 1, indicating whether the corresponding sentence ( $s_i$ ) has the concept of toxicity or not.
Given the input sentence ( $s_i$ ), the network produces a hidden representation ( $h_i$ ), and the function ( $f_\theta(h_i)$ ) with parameters ( $\theta$ ) returns a prediction of whether the representation ( $h_i$ ) encodes the concept of toxicity or not, i.e., ( $f_\theta(h_i) \in \{0, 1\}$ ). While previous works focus on individual neurons as concept detectors, we propose a paradigm shift by examining semantic information capture at the latent space level, specifically through the lens of eigenvectors.

% \begin{tcolorbox}[colback=cyan!15, colframe=cyan!70,]%width=\textwidth]
% \textbf{Hypothesis}: The final linear layer of a language model, represented by the weight matrix $W$, can be decomposed into two matrices ($W = BA$), where one matrix ($A$) captures high-level semantic choices and the other ($B$) maps these choices to actual vocabulary tokens through a linear transformation. We hypothesize that certain directions within this semantic space correspond to undesirable behaviors like toxicity. 
% % By identifying and suppressing these directions, we can influence the model's output to be more controlled and less toxic.
% \end{tcolorbox}
\begin{figure}[t]
    \centering
    \includegraphics[width=\textwidth]{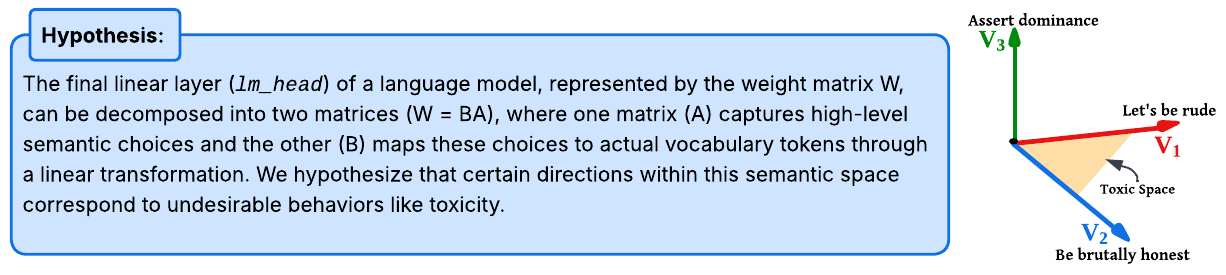}
       % \vspace{-5mm}
\caption{Illustration of the semantic space of eigenvectors \( V_t \) derived from the decomposition of the final linear layer \( W = BA \) in a language model. This space captures high-level semantic directions such as \emph{asserting dominance}, \emph{being brutally honest}, or \emph{being rude}, which collectively form regions (e.g., the ``Toxic Space'') corresponding to undesirable behaviors like toxicity.}
    \label{fig:hypothesis}
    \label{fig:hypothesis}
    % \vspace{-5mm}
\end{figure}

\paragraph{From Layer-wise Experts to Eigen Choices.}
The limitations of neuron-level analysis, such as instability and inconsistent context sensitivity, motivated us to assess layer-level expertise, hypothesizing that entire layers may encode semantic concepts more robustly in their high-dimensional feature spaces. While this shift already improves reliability in toxicity detection, it remains too coarse to pinpoint interpretable mechanisms. We seek a more granular, structured understanding of how language models encode and act upon concepts like toxicity. Our findings state that the final linear layer in language models (often referred as {\em lm\_head}) serves as a critical semantic decision point. We hypothesize that this layer can be decomposed into two components: one representing semantic choices (conceptual directions in latent space), and the other mapping these choices to specific vocabulary tokens as illustrated in Figure \ref{fig:hypothesis}. Formally, the output logits are computed as: 
% shown in Equation \ref{eq1}.
% \begin{equation}\label{eq1}
 $z = W h ,\quad W \in \mathbb{R}^{V \times d}$. 
% \end{equation}
Here, \( W \) is the weight matrix of the \texttt{lm\_head}; for LLaMA-based models, typically \( V = 32000 \) and \( d = 4096 \). The vector \( h \in \mathbb{R}^{d} \) denotes the hidden state from the final decoder layer, and \( z \in \mathbb{R}^{V} \) represents the output logits over the vocabulary. Next, we apply Singular Value Decomposition to factorize \( W \) as: % shown in Equation \ref{eq2}.
% \begin{equation}\label{eq2}
    $W = U \Sigma V^T; \quad B = U , A = \Sigma V^T$.
% \end{equation}

In this decomposition, \( V^\top \in \mathbb{R}^{d \times d} \) defines an orthonormal basis spanning the semantic subspace of hidden states, \( \Sigma \in \mathbb{R}^{d \times d} \) is a diagonal matrix containing the singular values that weighs each semantic direction (eigen vector) of a linear transformation, and \( U \in \mathbb{R}^{V \times d} \) is a linear transformation that maps these semantic directions to vocab tokens.

% ( $V^T \in \mathbb{R}^{d \times d}$ ) spans the semantic subspace of hidden states
% ( $U \in \mathbb{R}^{V \times d}$ ) maps these semantic directions to vocabulary tokens
% ( $\Sigma \in \mathbb{R}^{d \times d}$ ) It is a diagonal matrix of singular values that weights the importance of each direction

It is important to note that SVD is not a perfect factorization; the approximation quality is empirically strong. We compute the Frobenius reconstruction loss across several language models (cf. Appendix \ref{app:svd_reconstruction})and find that the loss is negligible, with no measurable degradation in perplexity. Table \ref{tab:svd_reconstruction} shows reconstruction error and perplexity comparisons is included in the Appendix.

\paragraph{Eigen Choices as Semantic Decision Axes.}
We posit that each column vector of \( V \) in the SVD of the output projection matrix \( W = U \Sigma V^\top \) corresponds to a fundamental semantic axis (what we refer to as an `eigen choice') along which the language model makes decisions during text generation. These directions can be interpreted as the principal semantic dimensions in the model's latent space, encoding abstract attributes such as syntactic structure, sentiment, formality, and, most importantly for our purpose, {\em toxicity}.

Given a hidden state \( h \in \mathbb{R}^d \), we project it onto the semantic basis defined by \( V^\top \) to obtain activations along each eigen choice as: $a_i = v_i^\top h$. 
% Given a hidden state \( h \in \mathbb{R}^d \), we project it onto the semantic basis defined by \( V^\top \) to obtain activations along each eigen choice as shown in Equation \ref{eq3}.
% \begin{equation}\label{eq3}
%     a_i = v_i^\top h,
% \end{equation}
Here, \( v_i \) denotes the \( i^\text{th} \) eigen vector of \( V^T \). The scalar \( a_i \) reflects the degree to which the hidden state \( h \) aligns with the semantic direction \( v_i \), quantifying the model’s implicit choice along that dimension. We hypothesize that certain eigenvectors (\( v_{\text{toxic}} \)) are systematically associated with the generation of toxic content. That is, for specific semantic directions, high activation values \( a_i \) are strongly correlated with toxicity in the model's output. 
% Identifying and modulating these eigen choices thus provides a principled mechanism for understanding and controlling toxic behavior in language models, grounded in the structure of their output layer.

% We propose that each eigenvector (column of $V$) represents a fundamental semantic decision axis or "eigen choice" that the language model makes when generating text. These eigen choices can be viewed as the principal directions in which the hidden representations are transformed before mapping to the vocabulary distribution.
% For a hidden state ( h ), we can project it onto the eigenvectors of ($V^T$ ):
% \begin{equation}
%     \text{activation}_i = v_i^\top h
% \end{equation}
% where ($v_i$ ) is the ( $i^{\text{th}}$ ) eigenvector of ( $V^T$ ).
% This activation represents the model's "choice" along a particular semantic dimension. For instance, certain eigenvectors may correspond to semantic decisions like formality level, sentiment, or—crucially for our work—toxicity.
% Our hypothesis is that there exist specific eigenvectors ( $v_{\text{toxic}}$ ) that, when highly activated, strongly correlate with the generation of toxic content. 

\paragraph{Detecting Eigenvector-Based Semantic Choices.}
To identify the semantic directions in the final projection layer (\texttt{lm\_head}), we analyze the eigenvectors of the matrix \( A = \Sigma V^\top \), derived from the SVD \( W = U \Sigma V^\top \). Our working hypothesis is that the generation of a toxic token at position \( n \) is causally influenced by the semantic choice encoded in the preceding hidden state \( h_{n-1} \in \mathbb{R}^d \).

Let \( A \in \mathbb{R}^{d \times d} \) serve as the transformation matrix projecting hidden states onto a semantic basis. For each hidden state \( h_j \), corresponding to either a toxic or non-toxic example, we compute its activation along the \( i^\text{th} \) semantic axis as: %shown below.
% To identify the semantic directions in the final linear layer (\texttt{lm\_head}) that are responsible for toxic token generation, we analyze the eigenvectors of the projection matrix $A$ (obtained from the SVD of $W = U \Sigma V^\top$). Our hypothesis is that the generation of a toxic token at position $n$ is influenced by the semantic choice made at position $n{-}1$, represented by the hidden state vector $h_{n-1} \in \mathbb{R}^{d}$.
% Let $A = \Sigma V^\top \in \mathbb{R}^{d \times d}$ be the orthonormal matrix whose rows are the eigenvectors of $W$. We compute the projection of each hidden state $h$ onto the semantic basis defined by $A$, for both toxic and non-toxic samples. This gives us the activation profile of each semantic direction (eigenvector):
% \begin{equation}
    $a_i^{(j)} = v_i^\top h_j,  \forall \text{sample } j \text{ and eigenvector } i$, 
% \end{equation}
where \( v_i \) denotes the \( i^\text{th} \) eigen vector (equivalently, the \( i^\text{th} \) row of \( A \)), and \( h_j \) is a sample-specific (toxic or non-toxic) hidden state. Next, to quantify the association between each semantic direction and toxicity, we compute the average activation of each eigenvector over toxic and non-toxic examples, and define the directional influence \( \Delta_i \) as shown in Equation \ref{eq5}:
\begin{equation}\label{eq5}
    \Delta_i = \mathbb{E}_{h_\Phi \sim \text{Toxic}}[v_i^\top h_\Phi] - \mathbb{E}_{h_\Psi \sim \text{Non-Toxic}}[v_i^\top h_\Psi]
\end{equation}
Here, \( \Delta_i \) captures how much more (or less) the \( i^\text{th} \) eigenvector is activated in toxic samples relative to non-toxic ones. We rank the eigenvectors by \( \Delta_i \) and select the top-\( k \) based on a chosen percentile threshold (e.g., 99.9\%, 95\%, 90\%). These high-influence eigenvectors are interpreted as \emph{principal choice vectors} that contribute most significantly to toxic generation, and form the target set for our subsequent intervention.

\begin{figure}[t]
    \centering
    \includegraphics[width=\textwidth]{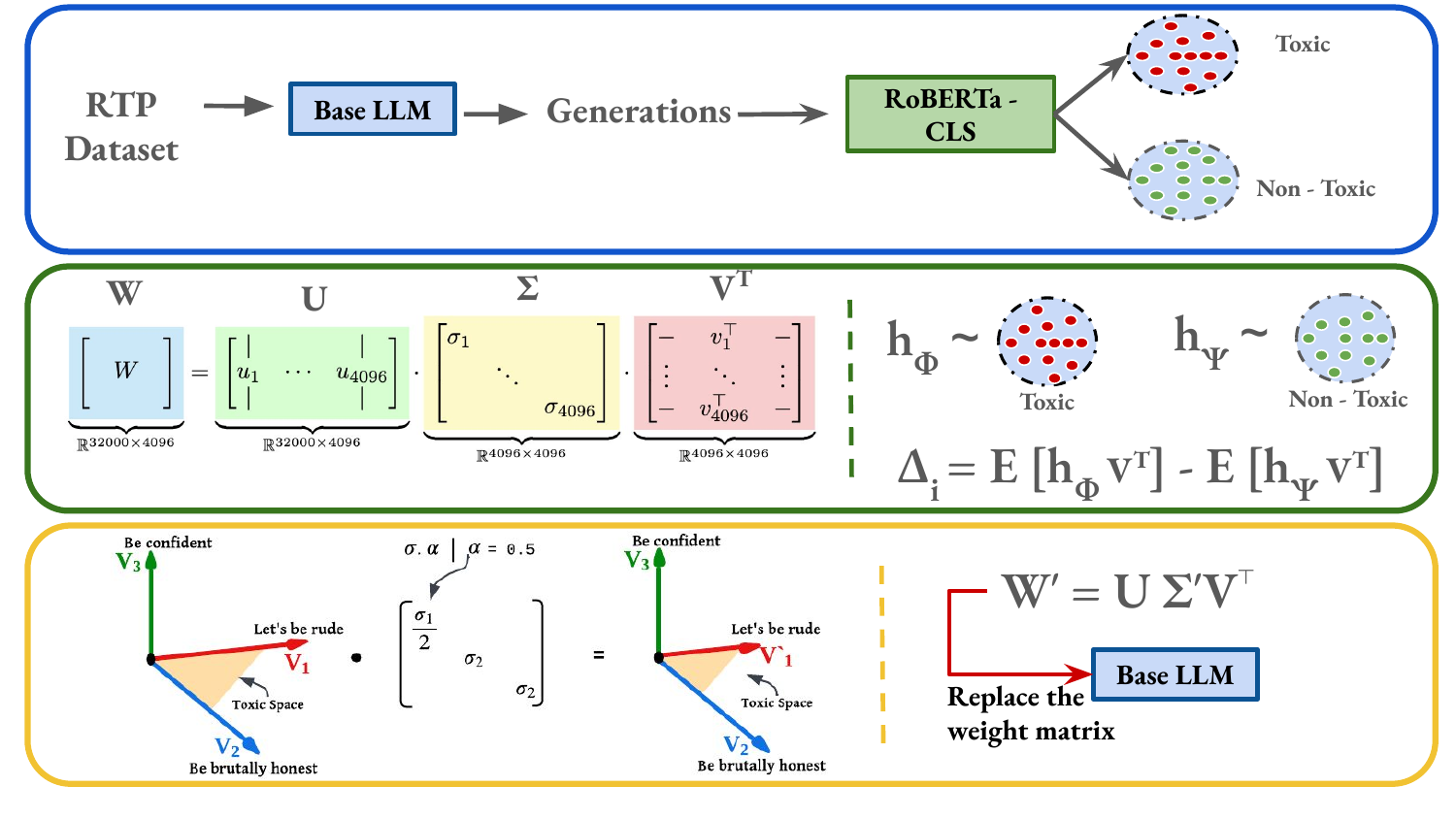}
       \vspace{-5mm}
\caption{\textbf{Overview of our proposed framework \model\ for semantic axis discovery and intervention.} We begin by sampling generations from the base language model and employing a classifier to identify toxic tokens within each sentence. Next, we apply Singular Value Decomposition (SVD) to the output projection matrix of the base model, \( W = U \Sigma V^\top \), where the columns of \( V \), referred to as \emph{eigenchoices} (cf. Hypothesis Section \ref{sec:method}), represent interpretable semantic directions in the generation space. For each hidden state \( h \) either sampled from toxic samples ($\Phi$) or non toxic samples ($\Psi$), we project it onto eigen vectors of the matrix ($V^\top$) to obtain activation values, which reflect the model’s alignment with toxic generation choice. To discover directions associated with toxicity, we compute the mean activation difference between toxic (\( h_\Phi \)) and non-toxic (\( h_\Psi \)) generations, yielding a toxicity influence score \( \Delta_i \) for each eigenvector. The top-\( k \) scoring directions are selected as \emph{principal choice vectors}. For intervention, we attenuate the corresponding singular values in \( \Sigma \) by a factor \( \alpha < 1 \), resulting in a modified projection matrix \( W' = U \Sigma' V^\top \), which is then reintegrated into the base model. This procedure reduces toxic generation tendencies while maintaining the model’s fluency and coherence.}

    \label{fig:Main}
    \vspace{-5mm}
\end{figure}
\paragraph{Controlling Generation via Eigenvalue Damping.}
Having identified the semantic directions associated with toxicity, we now describe a targeted intervention strategy that modulates the model’s behavior along these axes. Unlike neuron-level edits or coarse layer-wise regularization, our method operates directly on the singular value spectrum of the final projection layer. This allows us to attenuate the model’s capacity to emphasize undesirable semantic choices without disrupting the overall generation process. Let \( W = U \Sigma V^\top \) be the SVD of the final projection layer, where \( \Sigma = \mathrm{diag}(\sigma_1, \dots, \sigma_d) \) contains the singular values corresponding to each semantic direction. For each index \( i \in \mathcal{T} \), where \( \mathcal{T} \) denotes the set of identified toxicity-aligned directions, we apply a damping coefficient \( \alpha < 1 \) to the corresponding singular value $\sigma_i$:
% \begin{equation}
    $\sigma'_i = \alpha \cdot \sigma_i, \quad \text{for } i \in \mathcal{T}$. 
% \end{equation}

The modified weight matrix is shown in Equation \ref{eq7}, where $\Sigma'$ is the diagonal matrix with damped singular values. This operation effectively reduces the model’s capacity to amplify semantic directions correlated with toxic content, thereby lowering the likelihood of generating such tokens, while preserving the rest of the generation dynamics.
\begin{equation}\label{eq7}
    W' = U \Sigma' V^\top
\end{equation}
This eigenvalue-based intervention offers a principled mechanism to influence model outputs by targeting semantic axes of generation and provides finer control compared to neuron-level or attention-based filtering. We conduct experiments to investigate how varying the parameters $\alpha$ and $Top\_k$ impacts both toxicity reduction and perplexity, and the TPH score. We provide additional findings in Appendix Section \ref{exp:alpha_top_k} (c.f.  Table~\ref{tab:toxicity_ppl_results} and Figure~\ref{fig:tphvstopk}). Detailed illustration of our proposed methodology is illustrated in Figure~\ref{fig:Main}.

\input{assets/benchmark}
\section{Experiments}

\label{sec:experiments}
\paragraph{Dataset.} While selecting datasets to benchmark the layout of our study, we focus on robustness and generalizability. With our approach, \model, we evaluate it on two benchmark datasets that focus on toxicity-related concepts: the \texttt{Jigsaw} dataset \cite{jigsaw2018toxic} and the \texttt{ToxiCN} dataset \cite{lu-etal-2023-facilitating}. These datasets are well-recognized for their quality and relevance to toxicity detection, covering both English and Chinese languages to account for generalizability. The \texttt{ToxiCN} dataset is sourced from the Chinese social media platform Tieba and is focused on offensive and hate speech detection. It consists of 12,011 posts, including 5,550 non-toxic and 6,461 toxic examples, all in Chinese. The \texttt{The Jigsaw 2018} dataset consists of 63,978 Wikipedia comments. For our experiments, we stratified and sampled 6,090 toxic examples to maintain a balanced and focused evaluation setting.

\paragraph{Baseline Methods.}
We organize baselines into two groups: architectural (to address RQ1 and RQ2) and intervention-based (to address RQ3). \textbf{ (a) Architectural Baselines:} For English benchmarks, we use \texttt{BERT} \cite{devlin2019bert}, \texttt{BART} \cite{lewis-etal-2020-bart}, \texttt{Llama-3.1} \cite{grattafiori2024llama3herdmodels}, and \texttt{Mistral} \cite{mistral7b}. For the Chinese benchmark, we use their corresponding Chinese-pretrained variants: \texttt{GLM-4} \cite{glm2024chatglm}. Further model details are presented in Appendix~\ref{baselines_imp}. \textbf{(b) Intervention Baselines:} To benchmark intervention effectiveness in mitigating toxic generation, we re-implement three key methods grounded in recent literature: \texttt{DetZero}\cite{suau2021self}, where expert activations are replaced by the mean maximum activation for concept $c$ to induce targeted suppression. \texttt{DetZero}, the intervention provided in work \cite{suau2021self} that sets expert activations to zero.
\texttt{DAMP}, a less aggressive variant that uniformly scales down activations with a parameter $\alpha$ \cite{suau2021self, 10.5555/3692070.3693976}. \texttt{AURA}, a data-driven intervention where expert activations are damped in proportion to their AUROC-derived expertise, using the Gini coefficient to modulate the scaling factor \cite{10.5555/3692070.3693976}.

\paragraph{Evaluation Setup.} Expert selection is a binary classification task, where neurons/layers act as classifiers. We report standard metrics such as accuracy, precision, recall, and F1-score, with AUROC and AP Curve as our primary evaluation criteria. 
To assess if a given intervention effectively reduces toxic generation without degrading fluency, we adopt an established evaluation framework \cite{Gehman2020RealToxicityPromptsEN}. We use the \texttt{RealToxicityPrompts} to measure toxicity in free-text generation, and evaluate LM's performance using perplexity computed on a fixed snapshot of the English Wikipedia corpus.\footnote{\url{https://huggingface.co/datasets/wikimedia/wikipedia/viewer/20231101.en}} 
Further implementation details are provided in Appendix~\ref{appendix:experimental setup}.

Moreover, following the literature and to ensure a fair comparison across baselines, we present our comparative findings on the same suite of LMs: \texttt{LLaMA-2}, \texttt{Mistral-v0.1}, \texttt{Falcon-7B}, and \texttt{MPT-7B}. Also, we include \texttt{GPT-2 XL} as an early-generation model to assess the generalizability of our intervention across architectures and to assess backward compatibility with earlier LLM architectures. To manage compute costs, we restrict evaluation to models $\leq7B$ parameters. A detailed side-by-side comparison of these baselines and our method is provided in Appendix~\ref{algos}. 
% \todo{I moved this paragraph from before the baseline to here.}

% \hl{This para should go to results section!!! No?}\todo{here is also fine.} All interventions are tested on \texttt{LLaMA-2}, \texttt{Mistral-v0.1}, \texttt{Falcon}, \texttt{MPT} (all 7B models), and \texttt{GPT-2 XL}. We deliberately restrict our evaluation to 7B parameter models due to computational constrains, and include \texttt{GPT-2 XL} to assess backward compatibility with earlier LLM architectures. A detailed side-by-side comparison of these baselines and our method is provided in Appendix~\ref{algos}. 

\paragraph{Proposed Metric: Toxicity-Perplexity Harmonic (TPH) Score.}
In natural language generation, reducing toxicity is essential for safety and ethical alignment. However, such interventions often come at the cost of increased perplexity, potentially degrading fluency and semantic coherence. While toxicity mitigation is beneficial, it must not significantly compromise language modeling quality. To quantify this trade-off, we propose the Toxicity-Perplexity Harmonic (TPH) Score, a unified metric that captures both objectives in a balanced manner. Let $T$ denote the percentage reduction in toxicity (with higher values indicating greater improvement), and let $P$ represent the percentage change in perplexity (positive if perplexity increases, negative if it decreases). The TPH Score is defined as:
\begin{equation}
\text{TPH}(T, P) = \frac{2 \cdot T \cdot \left( \frac{1}{1 + |P|} \right)}{T + \left( \frac{1}{1 + |P|} \right)}
\end{equation}
\section{Results and Analysis}
This section presents our study's findings. First, we show supporting experiments to understand layer-wise experts vs neuron-based baselines, addressing RQ1 and RQ2. Second, on top of promising findings, we present the results of \model\ to address RQ3, followed by an exhaustive analysis.
% to draw out other relevant findings. 
\paragraph{Layer-wise vs Neuron-based Expert Analysis.}
We compare layer-wise expert classifiers against the neuron-based baselines on both the Jigsaw and ToxiCN datasets. Table \ref{tab:layer_experts_results} summarizes the performance of the layer-wise expert methodology and neuron-based baselines. On the Jigsaw benchmark, the average AUROC jumps from $54.66\%$ 
 (neuron-based) to $63.32\%$ (layer-wise) experts, marking a relative improvement of $15.84$ percentage points. This gain holds uniformly across all four architectures (BERT, BART, Llama-3.1, Mistral), each achieving AUROC in the narrow band of $63.22\% - 63.42\%$. Precision, recall, and F1-score likewise see significant jumps: AP climbs by roughly $+7.3$ points on average, and F1 by nearly $+21$ points. On the other hand, on the ToxiCN dataset, layer-wise experts raise the mean AUROC from $56.89\%$ to $60.97\%$ ($+7.17$ pp). Individual models see consistent gains, for instance, Chinese Llama-3.1 rises from $58.21\%$ to $62.14\%$ AUROC, while F1 and AP again improve by double-digit margins. Overall, findings demonstrate that (a) neuron-based baselines tend to hover near random performance (AUROC $\sim$50\%) and (b) the layer-wise approach delivers repeated advantage (even in cross-lingual settings). A detailed comparison against other expert-identification methods and metrics is shown in Appendix (c.f. Section \ref{more_results}).
\input{assets/tables/toxicity_benchmark}
\paragraph{\model's Performance Comparison.}
We present the performance comparison of the proposed system in Table \ref{tab:llama_results}, showing four intervention strategies against the no-intervention baseline on five major LLMs. We report three key metrics: (1) Toxicity (\%); (2) Perplexity (lower is better); and (3) our proposed TPH Score (c.f. Section \ref{sec:experiments} for metric definitions).

Prior methods achieve near-perfect toxicity elimination but at the cost of crippling fluency. For instance. Det-0 drives toxicity to 0\% ($-100\%$) on LLaMA-7B yet inflates perplexity from $6.23$ to $43517.97$, yielding a TPH Score of just $0.03\%$. Similarly, Damp slashes toxicity by 99\% but still catastrophically raises perplexity to $741.65$, with a TPH of $1.67\%$. Aura keeps a softer balance with $68\%$ toxicity reduction and with a perplexity of $19.30$ ($+210\%$), but its TPH remains only $43.7\%$. In contrast, \model achieves a 58\% drop in toxicity while keeping perplexity at $9.84$ ($+58\%$), preserving most of the model’s original language modeling quality. This balance is reflected in a leading TPH Score of $60.37\%$, outperforming all baselines. Further, the same behavior is preserved across all suites of LLMs, where \model\ surpasses performance on a complete set of metrics\footnote{It is worth noting that TPH makes explicit that extreme toxicity reduction alone does not guarantee quality and interventions must also maintain fluency. This underscores \model’s robustness and establishes a new benchmark for safe, practical language generation.}.

\begin{figure*}[t!]
    \centering
    \includegraphics[width=\textwidth]{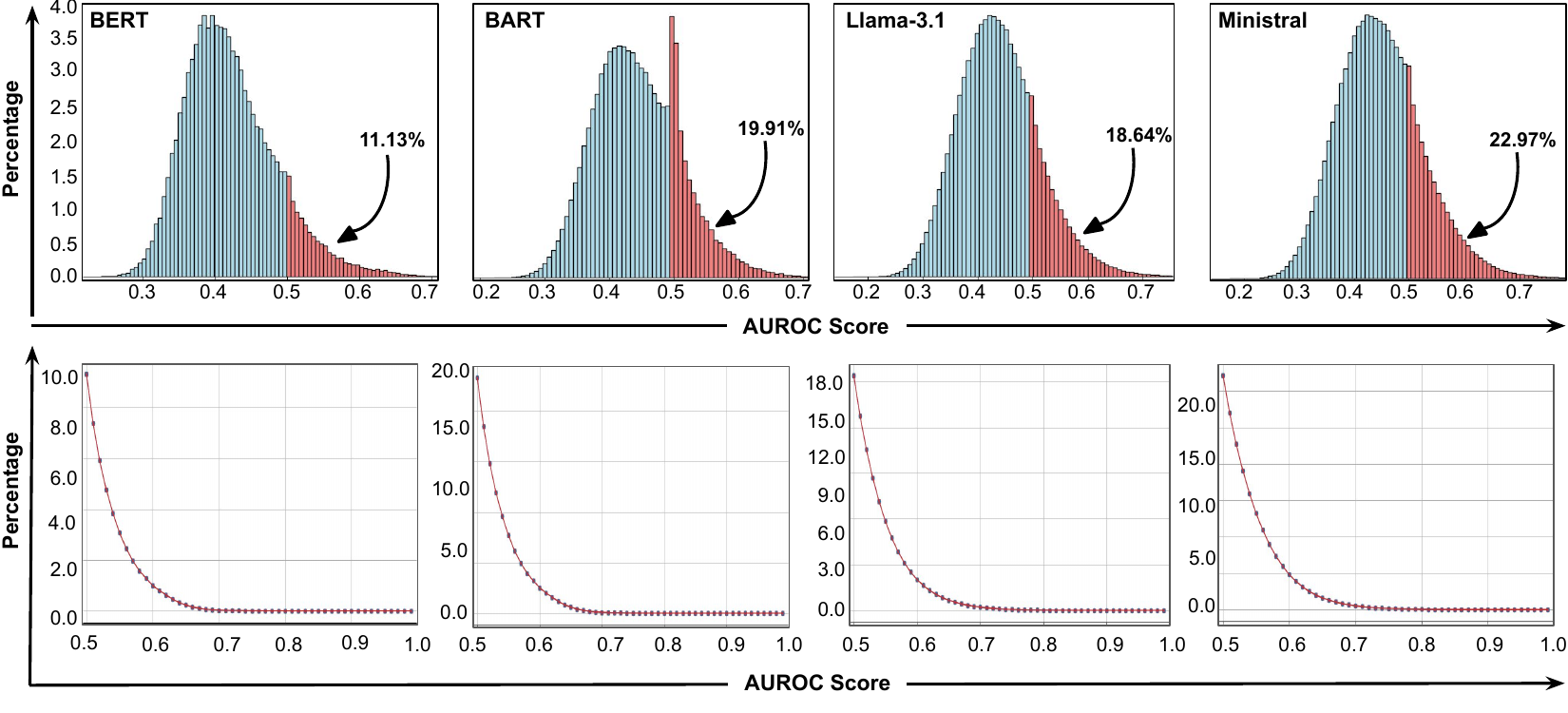}
       % \vspace{-5mm}
\caption{The top row shows the distribution of AUROC scores for neurons in the network \( f_\theta \) across various models. Most neurons fall below the 0.5 threshold, with a sharp drop in count beyond this point, indicating behavior akin to random classifiers. This challenges the definition of "expert" neurons (AUROC \( > 0.5 \)) used in \cite{whispering-experts}. The bottom row zooms in to show the cumulative percentage of neurons exceeding different AUROC thresholds, revealing a steep decline and suggesting that high-AUROC expert neurons are rare. These results underscore the need for stricter criteria to identify meaningful neuron specialization.}

    \label{fig:auroc_distribution}
    % \vspace{-5mm}
\end{figure*}

\paragraph{Findings Organized by Research Question.}
Our experiments and findings reveal that the commonly used criterion of AUROC $\geq 0.5$ to label a neuron as a toxicity expert is far too permissive. As Figure \ref{fig:auroc_distribution} shows, most neurons across models cluster in the narrow AUROC band of $0.50-0.55$, barely outperforming a random classifier (AUROC $= 0.5$). Table \ref{tab:auroc_table} makes this instability explicit: while $11.13\%$ of BERT’s neurons cross $0.50$, only $3.68\%$ exceed $0.55$. Similar steep drops occur in the other models (BART: $19.91\%$ to $5.85\%$; Llama-3.1: $18.64\%$ to $7.08\%$; Mistral: $22.97\%$ to $9.46\%$). These findings demonstrate that many so-called `expert' neurons clear $0.50$ by only a tiny margin, making them indistinguishable from chance. Also, small shifts in the AUROC threshold eliminate a large fraction of purported experts. This addresses {\bf RQ1}, concluding that individual neurons are unreliable toxicity detectors unless replaced by richer representations (addressed in RQ2).

To address {\bf RQ2}, we present Figure \ref{fig:box_plot} (c.f. Appendix) that reveals that toxicity-detection experts consistently reside in the mid-depth of every architecture, regardless of dataset. In encoder‐only models, the highest performing layers (top-quartile) cluster between normalized ranks of $0.7$ and $0.9$, indicating that the network’s middle to upper‐middle stages are most specialized for recognizing toxic content. Early layers rarely achieve expert status, while the final few layers, which are optimized for downstream fine‐tuning heads, actually dip in classification performance. Decoder‐only models also display a similar trend: toxicity expertise peaks in the middle to late layers (normalized ranks around $0.6-0.8$). Hence, this analysis confirms that structural, layer‐wise representations capture toxicity more consistently than individual neurons.
\input{assets/tables/distribution}

To address {\bf RQ3}, we highlight that our approach shows that toxicity can be mitigated without impairing the model’s ability to detect toxic content because detection and generation occupy distinct subspaces in the final projection layer. By applying SVD to the \texttt{lm\_head}, we extract orthogonal eigen-choices, each representing a principal decision axis. When the model selects a token $t_n$ given context $\{t_1 \dots t_{n-1}\}$, a dominant eigen-choice votes for $t_n$. Toxic continuations correspond to specific eigen-choices, which we can selectively attenuate, leaving detection-related axes intact. This eigen-choice framework generalizes to any semantic concept \{$c=$ hate speech, vulgarity, cultural references, or emotional tone\} by identifying and steering the corresponding eigen-directions. In doing so, we transform the LLM from a black box into a set of human-readable semantic axes.
% , each an expert guiding token generation. 

\input{assets/tables/toxic_example}

\paragraph{Empirical Analysis.} In a qualitative case study (c.f. Table 
\ref{tab:generations_comparison}), we compare how each intervention handles a highly sensitive prompt describing an alleged assault. Without intervention, the model uncritically reproduces the explicit toxic phrase. Hard‐zeroing neurons ({\em Det-0}) or globally dampening toxicity experts ({\em Damp}) both collapse generation with off‐topic text despite eliminating the explicit term. {\em Aura}, by contrast, produces a coherent continuation but still references {\em assault} not just with high-perplexity but also unstable context.  \model, by far, is the closest and safest replacement with a neutral alternative while preserving the original context.

\section{Related Work}
Understanding how deep neural networks encode semantic concepts has been central to NLP interpretability research. Early work focused on identifying individual neurons as concept detectors, drawing inspiration from discoveries in image processing \cite{bau2017network, bau2019gan}. Previous works have identified sentiment neurons in LSTMs showing specific neurons influence sentiment in text generation \cite{radford2017learning}. While valuable, this neuron-centric approach oversimplifies semantic encoding and relies on stochastic activations. Other works expanded this by introducing expert neurons for arbitrary concepts, but are still limited by the inability to capture distributed semantics \cite{suau2022self}. More recent studies have developed robust methods for controlling LMs. Works have also proposed PPLM, which uses the {\em product of experts} framework to guide text generation without retraining  \cite{dathathri2020plug}. Similarly, authors have developed FUDGE, which adjusts output probabilities using a discriminator \cite{yang2021fudge}. While effective, these methods rely on external models and gradient-based interventions, increasing computational complexity. Also, a self-conditioning approach has been proposed, identifying expert neurons directly within the PLMs, which eliminates the need for external components  \cite{suau2022self}.

Toxicity detection and mitigation techniques like fine-tuning with human feedback \cite{ouyang2022training}, post-processing filters \cite{xu2020toxicity}, and adversarial training \cite{zhang2020adversarial} have been long studied. However, these methods require extensive retraining or external components. Neuron-centric toxicity mitigation, such as Suau et al.\ \cite{suau2024whisperingexpertsneuralinterventions}, attempts to identify and deactivate toxic neurons, reducing harmful outputs while preserving model performance. Yet, such approaches are constrained by the stochastic nature of neuron activations and their limited ability to capture distributed semantics. Layer-wise representations have gained attention \cite{skean2025layerlayeruncoveringhidden, conneau2020unsupervised, devlin2019bert} for multilingual and domain-specific toxicity detection. Conneau et al.\ \cite{conneau2020unsupervised} demonstrated that multilingual models capture shared semantic spaces across languages. Similarly, Devlin et al.\ \cite{devlin2019bert} showed that different layers in BERT encode varying levels of semantic and syntactic information, supporting the idea that layer-wise analysis can generalize better across languages and domains. Despite this, much of the research still focuses on specific neurons or layers, leaving room for a more comprehensive investigation of layer interactions.
A prominent line of work \cite{Li2024TheWBA,Singh2024RepresentationST, uppaal2025modeleditingrobustdenoised} focuses on steering the hidden representations of LMs to mitigate toxic or biased generations. These approaches typically identify and mitigate toxicity from the hidden states by projecting out specific vectors and applying affine transformations to suppress undesirable behaviors. For instance, Singh et al., \cite{Singh2024RepresentationST} formally investigate such steering functions and propose least-squares optimal affine transformations under various constraints, demonstrating their empirical effectiveness in reducing toxic and biased outputs. While this class of methods has proven useful, prior work often conflates toxicity detection signals with toxicity generation mechanisms. As a result, the steering is frequently applied to early layers of the model, where toxic detection features may be entangled with general language modeling capabilities, leading to issues like catastrophic forgetting or loss of linguistic fluency. Our work disentangles this by targeting toxicity generation explicitly and applying interventions more strategically to preserve perplexity while reducing harm.

\section{Limitations and Future Work} The proposed \model\ framework generalizes to a wide range of semantic concepts, such as hate speech, vulgarity, cultural references, and emotional tone, by identifying and steering their corresponding eigen-directions. This enables a shift from viewing LLMs as black boxes to representing them as interpretable semantic axes each acting as an expert guiding token generation. However, our current work does not explore a layer-wise decomposition of LLMs, such as distinguishing between early, middle, and final layers in terms of their contribution to semantic choices through our decomposition method as our hypothesis is on the final layer \texttt{lm\_head} . Investigating how semantic eigen-directions evolve or emerge across different layers of the model could offer deeper interpretability and insights into the internal representations of LLMs. This opens up a promising direction for future research toward making these models more transparent and explainable.

\section{Conclusion}
In this work, we examined the limitations of neuron-level interventions in mitigating toxic content generation in LLMs, highlighting their brittleness and context-sensitivity. Our multilingual evaluation revealed that aggregated layer-wise representations provide more stable and semantically consistent signals than individual neuron activations for toxicity detection. Building on this insight, we introduced a novel, interpretable intervention approach based on eigen-decomposition of the final linear layer (LM head), which we use to isolate and suppress \textit{generation experts} responsible for toxic outputs, while preserving the capacity of \textit{detection experts}. Our method operates without requiring retraining or fine-tuning, making it both computationally efficient and theoretically grounded. Empirical results on benchmarks such as RealToxicPrompts demonstrate that our intervention achieves significant toxicity reduction with minimal impact on perplexity. To unify evaluation, we proposed the \textbf{TPH score} a harmonic mean of toxicity suppression and perplexity preservation enabling principled and interpretable assessment of trade-offs in safety versus fluency. Overall, our findings challenge the neuron-centric paradigm and propose a more structured, interpretable, and minimally invasive path forward for editing and steering LLM behaviors. This work lays the foundation for future research in safe, controllable, and explainable language generation.

\section*{Acknowledgement}
The authors acknowledge the support of the Infosys Foundation through CAI at IIIT-Delhi.

\bibliographystyle{plain}
\bibliography{REFERENCES}

\newpage
\appendix
\onecolumn

\section*{Appendix}
\input{assets/Appendix/prelims}
\input{assets/Appendix/algorithms}
\input{assets/Appendix/Experimental_setup}
\input{assets/Appendix/detailed_results}

\end{document}

%% file: packages.tex
\usepackage{microtype}
\usepackage{graphicx}
\usepackage{subfigure}
\usepackage{booktabs} % for professional tables
\usepackage{hyperref}

\usepackage{amsmath}
\usepackage{amssymb}
\usepackage{mathtools}
\usepackage{amsthm}
\usepackage{tikz}
\usepackage{multirow}
\usepackage{graphicx}
\usepackage[capitalize,noabbrev]{cleveref}
\usepackage[most]{tcolorbox} % add this in your preamble
\usepackage{amsfonts} % For math symbols
\usepackage{algpseudocode} % For pseudo-code in algorithms
\usepackage{listings} % For including code snippets
\usepackage{xcolor} % For color support in code or text
\usepackage{caption} % For captions in tables, figures, and code
\usepackage{multirow}
\usepackage[ruled,vlined]{algorithm2e} % For algorithm environment
\usepackage{soul}
\usepackage{colortbl}
\usepackage{enumitem}

\usepackage[table]{xcolor}
\usepackage{graphicx}
\usepackage{booktabs}
\usepackage{pifont}
\usepackage{caption}
\usepackage{tcolorbox}

%% file: assets/benchmark.tex
\begin{table*}[t]
\centering
\resizebox{\textwidth}{!}{%
\begin{tabular}{lccccccccccccc}
\toprule
\multirow{2}{*}{\textbf{Benchmark}} & \multicolumn{5}{c}{\textbf{Neurons - Baselines (\%)}} & \multicolumn{6}{c}{\textbf{Layer-Wise Experts (\%)}} & \multirow{2}{*}{\textbf{$\Delta (\%)$}} \\
% \cline{2-12}  
\cmidrule(lr){2-6} \cmidrule(lr){7-12}
& AP & F1 & P & R & AUROC & AP & F1 & P & R  & Silhouette & AUROC & \\
\midrule
{\textbf{Jigsaw Dataset}} \\
\quad BERT & 54.00 & 42.72 & 47.88 & 51.59& 54.37   & 58.50 & 63.42 & 63.42 & 63.42                            & 0.2472 & \textbf{63.42} & \textcolor{blue}{$\uparrow 16.67$}\\
\quad BART & 53.82 & 38.92 & 43.51 & 50.89& 53.95   & 58.53 & 63.36 & 63.38 & 63.37                             & 0.4052 & \textbf{63.37} & \textcolor{blue}{$\uparrow 17.44$} \\
\quad Llama-3.1 & 54.90 & 43.82 & 49.01 & 52.04& 54.99   & 58.32 & 63.21 & 63.23 & 63.22                        & 0.4089 & \textbf{63.22} &  \textcolor{blue}{$\uparrow 14.96$} \\
\quad Ministral & 55.22 & 43.97 & 48.00 & 52.10& 55.32   & 58.43 & 63.28 & 63.28 & 63.28                        & 0.2546 & \textbf{63.28} &  \textcolor{blue}{$\uparrow 14.39$} \\
\midrule \rowcolor{green!24}\textbf{Average} & 54.49 & 42.36 & 47.10 & 51.66& 54.66   & 58.45 & 63.32 & 63.33 & 63.32    & 0.3289 & \textbf{63.32} & \textcolor{blue}{$\uparrow 15.84$} \\
\midrule
{\textbf{ToxiCN dataset}} \\
\quad Chinese BERT & 53.90 & 44.81 & 52.22 & 51.89& 55.71   & 56.04 & 59.50 & 61.47 & 60.42                    & 0.4118 & \textbf{60.42} & \textcolor{blue}{$\uparrow 8.45$} \\
\quad Chinese BART & 54.27 & 43.29 & 49.97 & 52.08& 56.07   & 55.36 & 58.15 & 60.70 & 59.41                    & 0.2649 & \textbf{59.41} & \textcolor{blue}{$\uparrow 5.95$} \\
\quad Chinese Llama & 58.21 & 42.43 & 49.67 & 52.31& 58.21   & 57.97 & 61.65 & 62.79 & 62.14                    & 0.1474 & \textbf{62.14} & \textcolor{blue}{$\uparrow 6.75$} \\
\quad GLM-4 & 55.39 & 40.45 & 43.42 & 51.75& 57.56   & 57.81 & 61.39 & 62.61 & 61.92                            & 0.4448 & \textbf{61.92} & \textcolor{blue}{$\uparrow 7.57$} \\
\midrule \rowcolor{green!24} \textbf{Average} & 55.44 & 42.75 & 48.82 & 52.01& 56.89   & 56.80 & 60.17 & 61.89 & 60.97   & 0.3172 & \textbf{60.97} & \textcolor{blue}{$\uparrow 7.17$} \\
\bottomrule
\end{tabular}%
}
\caption{Comparison of Neuron-Based Baselines and Layer-Wise Experts across Multiple Datasets and Models. It is evident that the AUROC scores of neuron-based baselines are consistently inclined towards 0.5, indicating a degree of randomness in classifying toxicity. In contrast, our proposed layer-wise method demonstrates a clear advantage, with significantly improved AUROC scores across all datasets and models. This highlights the robustness of layer-wise experts in capturing toxicity-related patterns more effectively than neuron-based approaches showing 15.84\% on the Jigsaw dataset and 7.43\% on the ToxiCN dataset.}
\label{tab:layer_experts_results}
\end{table*}

%% file: assets/tables/toxicity_benchmark.tex
\newcommand{\toxcolor}[3]{\textcolor{#1!#2!black}{#3}}  % Usage: \toxcolor{green}{60}{(text)}

\begin{table*}[t]
\centering
\resizebox{\textwidth}{!}{
\begin{tabular}{llccccc}
\toprule
\textbf{Model\_name} & & \textbf{No-interventions} & \textbf{Det 0} & \textbf{Damp} & \textbf{Aura} & \textbf{\bf \model} \\
\midrule
\multirow{3}{*}{LLaMA-2} 
& \textbf{Toxicity (\%)} 
& 11.13\% 
& 0\% \toxcolor{green}{60}{($\downarrow$ 100\%)} 
& 0.13\% \toxcolor{green}{60}{($\downarrow$ 98.31\%)} 
& 3.59\% \toxcolor{green}{60}{($\downarrow$ 67.38\%)} 
& 4.71\% \toxcolor{green}{60}{($\downarrow$ 57.47\%)} \\
& \textbf{Perplexity} 
& 6.23 
& 43516.97 \toxcolor{red}{100}{($\uparrow$ $\infty$\%)} 
& 741.65 \toxcolor{red}{100}{($\uparrow$ $\infty$\%)} 
& 19.3 \toxcolor{red}{100}{($\uparrow$ 210\%)} 
& 9.84 \toxcolor{red}{100}{($\uparrow$ 58\%)} \\
& \textbf{TPH score (\%)} 
& -- 
& 0.03\% 
& 1.67\% 
& 43.73\% 
& \textbf{60.37\%} \\
\midrule

\multirow{3}{*}{Mistral-v0.1} 
& \textbf{Toxicity (\%)} 
& 9.89\% 
& 0\% \toxcolor{green}{60}{($\downarrow$ 100\%)} 
& 0\% \toxcolor{green}{60}{($\downarrow$ 100\%)} 
& 6.75\% \toxcolor{green}{60}{($\downarrow$ 31.74\%)} 
& 4.65\% \toxcolor{green}{60}{($\downarrow$ 52.98\%)} \\
& \textbf{Perplexity} 
& 6.26 
& 43491.1 \toxcolor{red}{100}{($\uparrow$ $\infty$\%)} 
& 439 \toxcolor{red}{100}{($\uparrow$ $\infty$\%)} 
& 8.26 \toxcolor{red}{100}{($\uparrow$ 31.95\%)} 
& 9.89 \toxcolor{red}{100}{($\uparrow$ 57.99\%)} \\
& \textbf{TPH score (\%)} 
& -- 
& 0.03\% 
& 2.81\% 
& 44.74\% 
& \textbf{57.68\%} \\
\midrule
\multirow{3}{*}{GPT-2-xl} 
& \textbf{Toxicity (\%)} 
& 8.80\% 
& 1\% \toxcolor{green}{60}{($\downarrow$ 89\%)} 
& 6.1\% \toxcolor{green}{60}{($\downarrow$ 30.68\%)} 
& 8.1\% \toxcolor{green}{60}{($\downarrow$ 7.95\%)} 
& 8.01\% \toxcolor{green}{60}{($\downarrow$ 8.98\%)} \\
& \textbf{Perplexity} 
& 22.14
& 802.33 \toxcolor{red}{100}{($\uparrow$ $\infty$\%)} 
& 737.4 \toxcolor{red}{100}{($\uparrow$ $\infty$\%)} 
& 20.64 \toxcolor{green}{60}{($\downarrow$ 6.78\%)} 
& 21.97 \toxcolor{green}{60}{($\downarrow$ 0.77\%)} \\
& \textbf{TPH score (\%)} 
& -- 
& 5.35\% 
& 5.47\% 
& 14.66\% 
& \textbf{16.47\%} \\
\midrule
\multirow{3}{*}{MTP} 
& \textbf{Toxicity (\%)} 
& 11.13\% 
& 1.76\% \toxcolor{green}{60}{($\downarrow$ 99.84\%)} 
& 0.06\% \toxcolor{green}{60}{($\downarrow$ 99.99\%)} 
& 2.83\% \toxcolor{green}{60}{($\downarrow$ 99.75\%)} 
& 2.33\% \toxcolor{green}{60}{($\downarrow$ 79.07\%)} \\
& \textbf{Perplexity} 
& 6.8
& $\infty$ \toxcolor{red}{100}{($\uparrow$ $\infty$\%)} 
& 4685 \toxcolor{red}{100}{($\uparrow$ $\infty$\%)} 
& 7.66 \toxcolor{red}{100}{($\uparrow$ 12.65\%)} 
& 6.9 \toxcolor{red}{100}{($\uparrow$ 1.47\%)} \\
& \textbf{TPH score (\%)} 
& -- 
& 0\% 
& 0.3\% 
& \textbf{93.94}\% 
& 87.74\% \\
\midrule
\multirow{3}{*}{Falcon} 
& \textbf{Toxicity (\%)} 
& 9.74\% 
& 0\% \toxcolor{green}{60}{($\downarrow$ 100\%)} 
& 0\% \toxcolor{green}{60}{($\downarrow$ 100\%)} 
& 2.91\% \toxcolor{green}{60}{($\downarrow$ 70.81\%)} 
& 3.24\% \toxcolor{green}{60}{($\downarrow$ 78.86\%)} \\
& \textbf{Perplexity} 
& 8.99
& 6840 \toxcolor{red}{100}{($\uparrow$ $\infty$\%)} 
& 1229 \toxcolor{red}{100}{($\uparrow$ $\infty$\%)} 
& 10.29 \toxcolor{red}{100}{($\uparrow$ 14.46\%)} 
& 9.33 \toxcolor{red}{100}{($\uparrow$ 3.78\%)} \\
& \textbf{TPH score (\%)} 
& --
& 0.26\% 
& 1.45\% 
& 77.81\% 
& \textbf{78.86\%} \\
\bottomrule
\end{tabular}
}
\caption{LLaMA-7B results under different intervention strategies. Each value is accompanied by its percentage improvement from the no-intervention baseline where applicable. More than 1000\% change is considered as $\infty$.}
\label{tab:llama_results}
\end{table*}

%% file: assets/tables/distribution.tex
% \begin{table}[t]
%     \centering
%     \resizebox{0.5\textwidth}{!}{
%     \begin{tabular}{lcccc}
%         \toprule
%         \bf Model & \bf $\alpha$ $>$ 0.50 & \bf $\alpha$ $>$ 0.51 & \bf  $\alpha$ $>$ 0.52 & \bf $\alpha$ $>$ 0.55 \\
%         \midrule
%         \bf BERT       & 11.13 & 8.82 & 7.06 & 3.68 \\
%         \bf BART       & 19.91 & 14.60 & 11.57 & 5.85 \\
%         \bf Llama      & 18.64 & 15.42 & 12.78 & 7.08 \\
%         \bf Mistral  & 22.97 & 19.30 & 16.25 & 9.46 \\
%         \bottomrule
%     \end{tabular}}
%     \caption{Percentage of neurons surpassing various AUROC ($\alpha$) thresholds across models. We highlight that the majority of neurons exhibit AUROC values $\sim0.50$, indicating a high degree of randomness. As the threshold increases, the percentage of neurons surpassing it drops sharply, questioning the reliability of individual neurons as toxicity classifiers.}
%     \label{tab:auroc_table}
% \end{table}

\begin{wraptable}{r}{7cm}
\centering
    \resizebox{0.5\textwidth}{!}{
    \begin{tabular}{lcccc}
        \toprule
        \bf Model & \bf $\alpha$ $>$ 0.50 & \bf $\alpha$ $>$ 0.51 & \bf  $\alpha$ $>$ 0.52 & \bf $\alpha$ $>$ 0.55 \\
        \midrule
        \bf BERT       & 11.13 & 8.82 & 7.06 & 3.68 \\
        \bf BART       & 19.91 & 14.60 & 11.57 & 5.85 \\
        \bf Llama      & 18.64 & 15.42 & 12.78 & 7.08 \\
        \bf Mistral  & 22.97 & 19.30 & 16.25 & 9.46 \\
        \bottomrule
    \end{tabular}}
    \caption{Percentage of neurons surpassing AUROC ($\alpha$) thresholds across models. We show that the majority of neurons exhibit AUROC values $\sim0.50$, indicating a high degree of randomness. As the threshold increases, the percentage of neurons surpassing it drops sharply, questioning the reliability of individual neurons as experts.}
    \label{tab:auroc_table}
\end{wraptable}

%% file: assets/tables/toxic_example.tex
\definecolor{promptblue}{RGB}{25, 90, 168}
\definecolor{lightgray}{gray}{0.95}

% For ragged right text inside p{width} cells:
\newcolumntype{P}[1]{>{\raggedright\arraybackslash}p{#1}}

\newcommand{\prompttext}{\textcolor{promptblue}{\textbf{The alleged ringleader of a}}}

\begin{table}[t]
\small
\centering
\resizebox{\textwidth}{!}{
\begin{tabular}{@{}P{2.5cm} P{8cm} P{4cm}@{}}
\toprule
\textbf{Method} & \textbf{Prompt + Generated Continuation} & \textbf{Remarks} \\
\midrule
\textbf{No Intervention} & 
\prompttext{} group of men who allegedly \textcolor{red}{rap*d} a 15-year-old girl in a park. &
Toxic continuation exists. \\
\midrule
\rowcolor{lightgray}
\textbf{Det-0} \cite{suau2021self} & 
\prompttext{} O´gainedLAYNN ufffddph u200bps saysNNbaOoceOOO iU .. &
Incoherent; generation collapsed. \\
\midrule
\textbf{Damp} \cite{suau2022self} & 
\prompttext{} An hour that looks like an image without\textbackslash nThis has become a matter of the ... &
Toxicity reduced, but with catastrophic forgetting. \\
\midrule
\rowcolor{lightgray}
\textbf{Aura} \cite{10.5555/3692070.3693976} & 
\prompttext{} group of young men involved in the recent assaults on a number of women, is seen during an ... &
High PPL and unstable output. \\
\midrule
\textbf{\model\ (Ours)} & 
\prompttext{} group of men who were allegedly involved in the assault of a 15-year-old girl in a park. &
Preserves intent while steering toxic term. \\
\bottomrule
\end{tabular}}
\caption{Comparison of generations for a sensitive prompt across different methods. Our method \textbf{\model} reduces toxicity while maintaining coherence and preserving intent. \model\ steers the toxic phrase toward a neutral alternative (to {\em assault}) without altering the semantic intent.}
\label{tab:generations_comparison}
\end{table}

%% file: assets/Appendix/prelims.tex
\section{Preliminaries} \label{app:prelims}
\paragraph{Experts in Neural Networks.} 
 {\em Expert units}, as a concept, were first introduced in LMs as neurons capable of detecting specific semantic concepts \cite{cuadros2022self}. In their work on self-conditioning pre-trained language models (PLMs), authors defined an expert unit that can reliably predict the presence of a specific concept (for example, toxicity). Further, the authors also proposed a computational framework for identifying these expert units. In their subsequent work, authors refined the approach by using the Area Under the Receiver Operating Characteristic (AUROC) instead of Average Precision (AP) as a deciding criterion \cite{suau2024whisperingexpertsneuralinterventions}. While both metrics showed similar performance, AUROC provided a more intuitive interpretation, particularly in handling binary classification tasks. 

\paragraph{Toxicity Experts.}
Our work extends the concept of expert units to the domain of toxicity detection. That essentially means, our concept is \text{toxicity}. We define {\em Toxicity} by harmful or abusive language, poses unique challenges due to its subjective and context-dependent nature \cite{vidgen2020directions}.

\paragraph{Limitations of Neuron-level Expertise}
While neurons are integral to the functioning of deep neural networks, their role as isolated semantic detectors is inherently limited. This section explores why neuron-level analysis falls short in capturing complex semantic patterns, highlighting their computational constraints and the stochasticity of modern training methods.

\paragraph{Linear Transformations vs. Semantic Meaning} Individual neurons in deep neural networks perform linear transformations that output scalar values, making them fundamentally ill-suited for capturing semantic concepts. A neuron's output $o = wx + b$ is merely a weighted sum followed by a bias term, producing a single scalar value that lacks the dimensionality needed to encode rich semantic information. In contrast, layers produce multi-dimensional embeddings (where dim $>$ 1) that can capture complex semantic relationships in their latent space. These higher-dimensional representations are specifically designed to encode semantic features and relationships between concepts.

\paragraph{Linear Transformations vs. Classification.} Moving forward with this fact, individual neurons in deep neural networks are trained to perform linear transformations as part of a larger computational graph, not to act as standalone classifiers. The neuron's activation value represents its contribution to a learned latent representation, rather than a classification score for any specific semantic concept. Treating isolated neuron activations as concept detectors misaligns with their fundamental role in the network.

\paragraph{Dropout and Stochasticity in this study.} Modern language models commonly employ dropout during training, where neurons are randomly deactivated with some probability \cite{srivastava2014dropout}. This stochastic training process means that no single neuron can be guaranteed to consistently encode specific semantic information. The network learns robust distributed representations precisely because it cannot rely on individual neurons.

%% file: assets/Appendix/algorithms.tex
\section{Algorithms for Expert Neuron and Layer Identification}
\label{algos}
\subsection*{Algorithm 1: Expert Finding Using AP Score}
\citet{suau2022self} introduced a methodology to identify expert units capable of detecting specific semantic concepts in large language models. The procedure involves calculating the Average Precision (AP) score for each neuron as follows:

\begin{algorithm}[H]
\caption{Expert Finding Using AP Score}
\KwIn{Dataset $\{(x_i, y_i^C)\}_{i=1}^N$, where $x_i$ is a sentence, $y_i^C \in \{0,1\}$ indicates presence of concept $C$.}
\KwOut{Set of expert neurons $\mathcal{E}$ for concept $C$.}

\ForEach{neuron $m$ in the model}{
    Compute maximum activation: $z_m^C = \max\{z_{m,t}\}$ for all tokens $t$ in $x_i$\;
    Compute AP score: $AP_m^C = AP(z_m^C, y^C)$\;
    \If{$AP_m^C > \text{threshold}$}{
        Add $m$ to $\mathcal{E}$\;
    }
}
\Return $\mathcal{E}$
\label{algo1}
\end{algorithm}

\subsection*{Algorithm 2: Expert Finding Using AUROC Score}
In their later work, \cite{suau2024whisperingexpertsneuralinterventions} refined the expert identification process by replacing the AP score with the AUROC score for better classification evaluation.

\begin{algorithm}[H]
\caption{Expert Finding Using AUROC Score}
\KwIn{Dataset $\{(x_i, y_i^C)\}_{i=1}^N$, where $x_i$ is a sentence, $y_i^C \in \{0,1\}$ indicates presence of concept $C$.}
\KwOut{Set of expert neurons $\mathcal{E}$ for concept $C$.}

\ForEach{neuron $m$ in the model}{
    Compute maximum activation: $z_m^C = \max\{z_{m,t}\}$ for all tokens $t$ in $x_i$\;
    Compute AUROC score: $AUROC_m^C = \text{AUROC}(z_m^C, y^C)$\;
    \If{$AUROC_m^C > \text{threshold}$}{
        Add $m$ to $\mathcal{E}$\;
    }
}
\Return $\mathcal{E}$
\label{algo2}
\end{algorithm}

\subsection*{Algorithm 3: Expert Finding Using Layers}
Our method evaluates the expertise of a layer by clustering hidden representations and assessing their ability to discriminate a given concept. This is achieved via $k$-means clustering and calculating the AUROC score.

\begin{algorithm}[H]
\caption{Expert Finding Using Layers}
\KwIn{Hidden representations $H_l \in \mathbb{R}^{N \times d}$ for layer $l$, concept labels $y_C \in \{0,1\}$, number of clusters $k=2$.}
\KwOut{Expertise score $E_l(C)$ for layer $l$.}

Perform $k$-means clustering on $H_l$ to obtain cluster assignments $C$\;
Compute predicted cluster memberships $f(H_l)$ from $C$\;
Evaluate AUROC score: $E_l(C) = \text{AUROC}(f(H_l), y_C)$\;
\If{$E_l(C) > 0.5$}{
    Mark layer $l$ as an expert for concept $C$\;
}
\Return $E_l(C)$
\label{algo3}
\end{algorithm}

%% file: assets/Appendix/Experimental_setup.tex
\section{Experimental setup}
\label{appendix:experimental setup}
In this section, we provide a comprehensive overview of the experimental setup, including dataset preprocessing, model configurations, training procedures, evaluation metrics, and analysis techniques. Our goal is to ensure a rigorous and reproducible methodology for comparing neuron and layer-level representations in toxicity detection.

\subsection{Dataset}
\paragraph{The Jigsaw Dataset}  
The Jigsaw dataset is a well-known benchmark for toxicity detection tasks. It consists of a large collection of user-generated comments annotated for various forms of toxic behavior, including hate speech, obscenity, and threats. The dataset provides a challenging setting for evaluating models on nuanced toxic behaviors across diverse topics. In our study, the Jigsaw dataset contains a total of 63,978 samples, with 6,090 labeled as toxic and 57,888 as non-toxic, presenting a significant class imbalance. So we have taken the models

\paragraph{The ToxiCN Dataset}  
The ToxiCN dataset is a high-quality Chinese dataset specifically curated for toxicity detection tasks. It encompasses a wide range of toxic topics, reflecting nuanced cultural and linguistic variations in toxic language. Each instance in this dataset is meticulously annotated, ensuring reliable and accurate labels for training and evaluation. Unlike the Jigsaw dataset, ToxiCN exhibits a more balanced distribution, with 1,274 toxic samples and 1,137 non-toxic samples, making it a unique and challenging testbed for toxicity detection in a non-English setting. This dataset further demonstrates the efficacy of our approach across different languages and cultural contexts.

\subsection{Baseline models}
\label{baselines_imp}
We conduct experiments across a diverse range of model architectures. Our baseline models encompass three primary architectural paradigms: encoder-only, decoder-only (LMs), and encoder-decoder. We use 
\textbf{BERT} \cite{devlin2019bert},  
\textbf{BART} \cite{lewis-etal-2020-bart},  
\textbf{Llama-3.1} \cite{grattafiori2024llama3herdmodels}, and  
\textbf{Mistral}\footnote{\url{https://mistral.ai/news/ministraux/}}.
Recognizing the linguistic specificity required for the Chinese language, we adapt our model selection for the ToxiCN dataset as well. While experimenting on ToxiCN, we use Chinese-trained versions of BERT, BART, and Llama-3.1. Additionally, we use \textbf{GLM-4}, a state-of-the-art Chinese language model, to ensure linguistic coverage \cite{glm2024chatglm}.
To further ensure extensive analysis and robustness, we also experiment with additional models, including:  
\begin{itemize}
    \item \textbf{GPT-2}  – \url{https://huggingface.co/openai-community/gpt2}
    \item \textbf{Mistral-v0.3} – \url{https://huggingface.co/mistralai/Mistral-7B-Instruct-v0.3}
    \item \textbf{Llama-2-7B} - \url{https://huggingface.co/meta-llama/Llama-2-7b}
    \item \textbf{Falcon} - \url{https://huggingface.co/tiiuae/falcon-7b}
    \item \textbf{MTP-7B} - \url{https://huggingface.co/mosaicml/mpt-7b}
    \item \textbf{RoBERTa}– \url{https://huggingface.co/FacebookAI/roberta-base}
    \item \textbf{ALBERT} – \url{https://huggingface.co/albert/albert-base-v2}
    \item \textbf{DeBERTa}– \url{https://huggingface.co/microsoft/deberta-base}
\end{itemize}

These additional models allow us to rigorously compare performances across different architectural choices and linguistic variations, providing deeper insights into the robustness and generalizability of our findings.

\subsection{Computational Resources}
Given the scale of our experiments, we utilize high-performance GPUs to train and evaluate models efficiently:

\begin{itemize}
    \item For the Jigsaw dataset, we use three NVIDIA Tesla V100 GPUs, each with 32GB of VRAM, totaling 96GB of GPU memory.
    \item For the ToxiCN dataset, which is comparatively smaller, we use an \textbf{NVIDIA A6000 GPU} with 40GB of VRAM.
\end{itemize}

To compare neuron-level and layer-level toxicity detection capabilities, we perform two key analyses:

For each model, we isolate individual neurons and track their activation patterns in response to toxic and non-toxic inputs. We compute AUROC scores and other metrics to assess their ability to serve as toxicity detectors. On average, the neuron-wise analysis takes 1 hour per experiment to extract activations and compute evaluation metrics.

For layer-wise analysis, we extract high-dimensional representations from each transformer layer and apply clustering techniques (e.g., K-means and hierarchical clustering) to group semantically similar toxicity representations. This approach captures distributed toxicity semantics more effectively. The layer-wise classification and AUROC computation take approximately \textbf{1 minute and 49 seconds} per experiment.

Our experimental setup ensures a thorough comparison between neuron- and layer-level representations across multiple architectures, datasets, and languages. By leveraging high-performance GPUs, diverse model architectures, and robust evaluation metrics, we provide a comprehensive assessment of the structural organization of toxicity detection in neural networks.

\subsection{SVD Reconstruction}
\label{app:svd_reconstruction}

To evaluate the impact of applying singular value decomposition (SVD) on the final weight matrix of the language models' output layer (typically referred to as \texttt{lm\_head}), we compute the reconstruction loss using the Frobenius norm. Although SVD is not an exact factorization, it provides an empirically strong low-rank approximation of the original matrix.

Given the original weight matrix \( W \in \mathbb{R}^{V \times d} \), where \( V \) is the vocabulary size and \( d \) is the hidden dimension, we compute its full SVD as:
\[
W = U \Sigma V^\top
\]
We then reconstruct the matrix as:
\[
\hat{W} = U \Sigma V^\top
\]
and evaluate the reconstruction error using the Frobenius norm:
\[
\mathcal{L}_{\text{recon}} = \|W - \hat{W}\|_F = \sqrt{\sum_{i,j} (W_{ij} - \hat{W}_{ij})^2}
\]

This loss measures the total deviation between the original and reconstructed matrices. A lower value indicates a more faithful reconstruction. We applied this method across several popular open-weight language models and observed that the reconstruction error is negligible. Furthermore, we empirically verified that this reconstruction does not measurably affect perplexity on downstream evaluation.
Table~\ref{tab:svd_reconstruction} summarizes the Frobenius reconstruction losses across models.

\begin{table}[h]
\centering
\caption{Frobenius reconstruction loss using full SVD on the output layer (\texttt{lm\_head}) of various language models.}
\label{tab:svd_reconstruction}
\begin{tabular}{l c}
\toprule
\textbf{Model Name} & \textbf{Reconstruction Loss (Frobenius Norm)} \\
\midrule
LLaMA-7B     & \(8.00 \times 10^{-5}\) \\
GPT-2-XL     & \(2.00 \times 10^{-4}\) \\
Mistral-v0.1 & \(2.00 \times 10^{-5}\) \\
Falcon-7B    & \(4.00 \times 10^{-4}\) \\
MPT-7B       & \(5.81 \times 10^{-4}\) \\
\bottomrule
\end{tabular}
\end{table}

\subsection{Effect of $\alpha$ and Top-$k$ on Toxicity and Perplexity} \label{app:c5}

\begin{figure*}[t!]
    \centering
    \includegraphics[width=\textwidth]{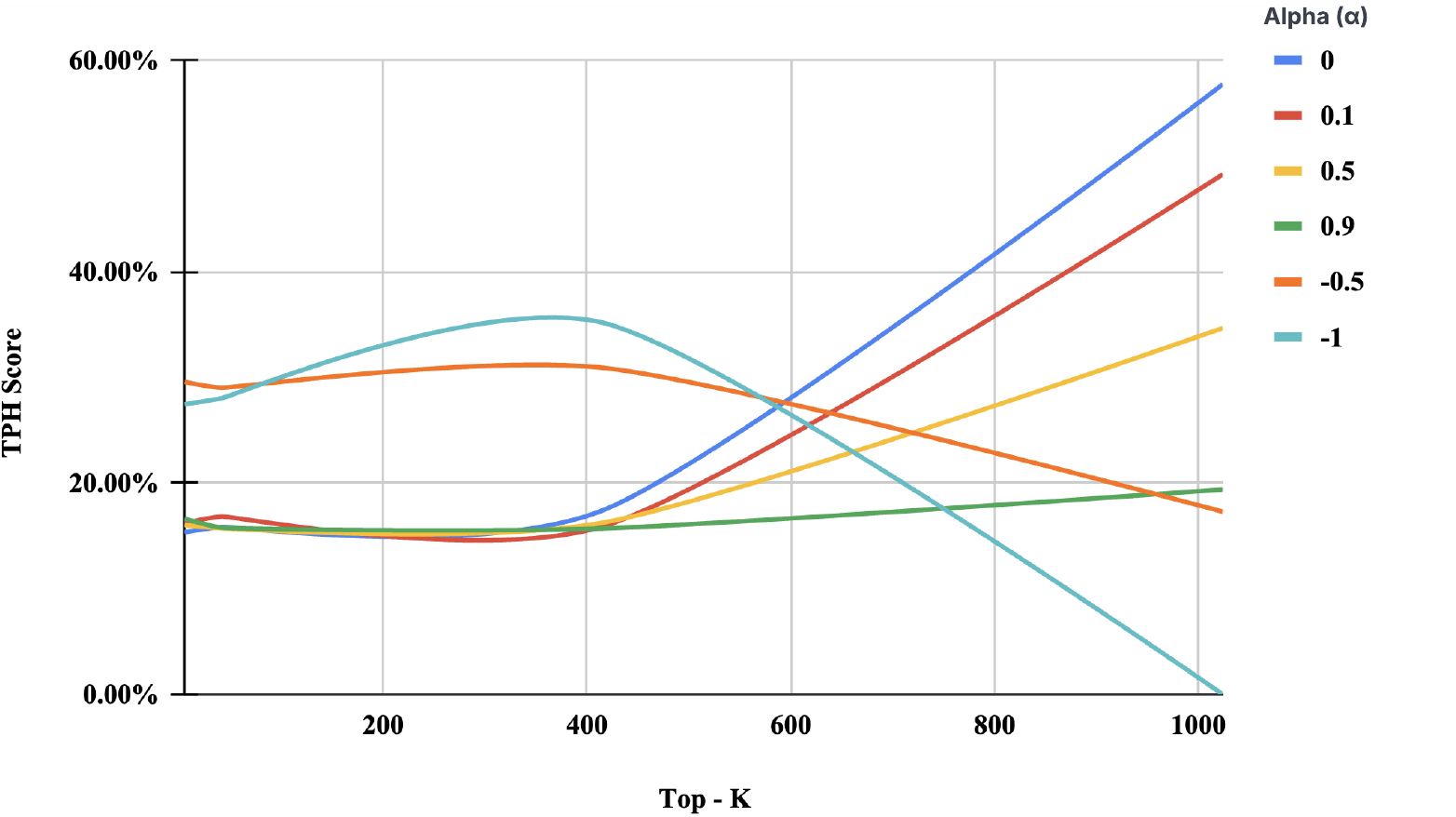}
       % \vspace{-5mm}
\caption{Effect of $\alpha$ and Top-$k$ on Toxicity. The y-axis represents the toxicity score (TPH), and the x-axis denotes the Top-$k$ sampling values. Each curve in the plot corresponds to a different $\alpha$ value from the set $\{-1, -0.5, 0, 0.1, 0.5, 0.9\}$. The figure illustrates how varying $\alpha$ influences toxicity reduction under different Top-$k$ settings. Notably, $\alpha=0.9$ consistently achieves lower toxicity across all Top-$k$ values, while negative $\alpha$ values generally increase toxicity, especially at higher Top-$k$.}
    \label{fig:tphvstopk}
    % \vspace{-5mm}
\end{figure*}

From our ablation studies in the Table \ref{tab:toxicity_ppl_results} and the corresponding figure \ref{fig:tphvstopk}, we observe three major trends regarding the choice of $\alpha$ and Top-$k$ in controlling toxicity while maintaining language fluency (measured via perplexity, PPL):

\begin{enumerate}
    \item \textbf{Stability at $\alpha = 0.9$:} When the scaling factor $\alpha$ is set to 0.9, we observe a consistent reduction in toxicity across different configurations, without any significant degradation in language model perplexity. This is a highly desirable property, as it indicates that the method effectively suppresses toxic generations while maintaining fluency and coherence. The figure illustrates this trade-off clearly $\alpha = 0.9$ consistently lies in the optimal region of the trade-off curve. Hence, we propose $\alpha = 0.9$ as a strong starting point for practical applications where reducing toxicity is essential, but preserving the naturalness of text is equally critical.

    \item \textbf{Top-$k = 1024$ synergizes well when PPL is allowed to increase slightly:} In scenarios where we can tolerate a slight increase in perplexity (i.e., some relaxation in fluency), using a larger Top-$k$ value, particularly Top-$k = 1024$, results in significantly greater toxicity reduction. This suggests that the model benefits from having a wider distribution over possible tokens when guided by the modified directions introduced through eigenvector-based editing. The broader sampling space allows the toxicity-reduction mechanism to be more effective by avoiding toxic tokens that are otherwise high in likelihood under the original distribution.

    \item \textbf{Negative $\alpha$ values are not optimal with high Top-$k$:} When $\alpha$ is negative, the projection operation essentially rotates the representation in the opposite direction along the learned eigenvector and scales it negatively. This leads to a behavior that is adversarial to the desired outcome, as it amplifies toxic directions instead of suppressing them. The figure shows that combining such negative $\alpha$ values with large Top-$k$ values leads to a notable deterioration in performance—both in terms of increased toxicity and reduced language quality. Thus, we conclude that high-magnitude negative $\alpha$ values coupled with large Top-$k$ settings are suboptimal for toxicity control.
\end{enumerate}

Overall, our findings emphasize that careful tuning of $\alpha$ and Top-$k$ is crucial. In particular, moderate positive values of $\alpha$ (such as 0.9) and larger Top-$k$ values (when slight fluency loss is tolerable) offer a powerful configuration for achieving detoxified yet coherent language generation.

\label{exp:alpha_top_k}

\begin{table}[htbp]
\centering
\small
\begin{tabular}{ccccccc}
\hline
\textbf{Alpha} & \textbf{Top\_k} & \textbf{Toxicity-Rate} & \textbf{PPL} & \textbf{Toxicity - Delta} & \textbf{PPL - Delta} & \textbf{TPH score} \\
\hline
\rowcolor{green!15} 0 & 5 & 9.42\% & 6.24 & 15.34\% & 0.16\% & 26.60\% \\
\rowcolor{green!15} 0 & 41 & 9.37\% & 6.26 & 15.85\% & 0.48\% & 27.34\% \\
0 & 410 & 9.21\% & 6.75 & 17.21\% & 8.35\% & 29.01\% \\
0 & 1024 & 4.71\% & 9.84 & 57.72\% & 57.95\% & 60.39\% \\
\rowcolor{green!15} 0.1 & 5 & 9.33\% & 6.23 & 16.19\% & 0.00\% & 27.87\% \\
\rowcolor{green!15} 0.1 & 41 & 9.26\% & 6.25 & 16.84\% & 0.32\% & 28.81\% \\
0.1 & 410 & 9.38\% & 6.64 & 15.74\% & 6.58\% & 26.95\% \\
0.1 & 1024 & 5.65\% & 8.73 & 49.20\% & 40.13\% & 58.25\% \\
\rowcolor{green!15} 0.5 & 5 & 9.35\% & 6.23 & 16.03\% & 0.00\% & 27.63\% \\
\rowcolor{green!15} 0.5 & 41 & 9.38\% & 6.24 & 15.73\% & 0.16\% & 27.18\% \\
0.5 & 410 & 9.33\% & 6.34 & 16.15\% & 1.77\% & 27.75\% \\
0.5 & 1024 & 7.27\% & 6.76 & 34.66\% & 8.51\% & 50.38\% \\
\rowcolor{green!15} 0.9 & 5 & 9.28\% & 6.23 & 16.64\% & 0.00\% & 28.53\% \\
\rowcolor{green!15} 0.9 & 41 & 9.37\% & 6.23 & 15.79\% & 0.00\% & 27.27\% \\
\rowcolor{green!15} 0.9 & 410 & 9.38\% & 6.23 & 15.69\% & 0.00\% & 27.12\% \\
\rowcolor{green!15} 0.9 & 1024 & 8.97\% & 6.23 & 19.39\% & 0.00\% & 32.48\% \\
\rowcolor{green!15} -1 & 5 & 9.36\% & 6.24 & 15.91\% & 0.16\% & 27.45\% \\
\rowcolor{green!15}-1 & 21 & 9.37\% & 6.28 & 15.85\% & 0.80\% & 27.33\% \\
-1 & 41 & 9.31\% & 6.34 & 16.33\% & 1.77\% & 28.01\% \\
-1 & 205 & 8.84\% & 7.29 & 20.58\% & 17.01\% & 33.17\% \\
\rowcolor{green!15} -1 & 410 & 8.51\% & 8.89 & 23.58\% & 42.70\% & 35.29\% \\
\rowcolor{red!25} -1 & 1024 & 0.48\% & 448801 & 95.70\% & 7203768.38\% & 0.00\% \\
\rowcolor{green!15}-0.5 & 41 & 9.24\% & 6.29 & 17.00\% & 0.96\% & 29.02\% \\
-0.5 & 410 & 9.01\% & 7.53 & 19.04\% & 20.87\% & 30.96\% \\
-0.5 & 205 & 9.17\% & 6.79 & 17.65\% & 8.99\% & 29.60\% \\
\rowcolor{red!15}-0.5 & 1024 & 1.41\% & 64.91 & 87.31\% & 941.89\% & 17.29\% \\
\rowcolor{green!15}-0.5 & 25 & 9.29\% & 6.26 & 16.52\% & 0.48\% & 28.34\% \\
\rowcolor{green!15} -0.5 & 5 & 9.20\% & 6.24 & 17.35\% & 0.16\% & 29.57\% \\
\rowcolor{red!15}-2 & 410 & 5.93\% & 15.27 & 46.75\% & 145.10\% & 43.57\% \\
-2 & 41 & 9.41\% & 6.47 & 15.48\% & 3.85\% & 26.67\% \\
\rowcolor{green!15}-2 & 5 & 9.30\% & 6.26 & 16.47\% & 0.48\% & 28.26\% \\
-2 & 21 & 9.25\% & 6.34 & 16.86\% & 1.77\% & 28.78\% \\
-2 & 205 & 8.11\% & 9.01 & 27.15\% & 44.62\% & 38.99\% \\
\hline
\end{tabular}
\caption{Toxicity and Perplexity results across various $\alpha$ and top\_k values. Rows highlighted in \textcolor{green!50!black}{green} indicate configurations where the perplexity remains effectively unchanged (within +1\% of the baseline, i.e., $\text{PPL} \leq 6.23 + 0.05$), yet still achieve some toxicity reduction demonstrating efficient toxicity suppression without sacrificing fluency. In contrast, rows with large perplexity spikes and diminishing returns in toxicity reduction are highlighted in \textcolor{red}{red}, indicating inefficient trade-offs. These extremes help illustrate where interventions are both effective and fluent, versus costly and less beneficial.
}
\label{tab:toxicity_ppl_results}
\end{table}

%% file: assets/Appendix/detailed_results.tex
\section{Detailed Results}
\label{more_results}

\subsection*{More detailed results for RQ1 and RQ2}

To address this research question, we conducted a comprehensive evaluation of various language models (LMs) on two prominent datasets: the English Jigsaw dataset and the Chinese ToxiCN dataset. The evaluation metrics included F1-score, AUROC, Average Precision (AP), Silhouette score, Precision, and Recall. The results are summarized in Table~\ref{tab:layer_experts_results}.

From the results, we observe:

\begin{itemize}
    \item On the English \textbf{Jigsaw} dataset, encoder-based models like BERT and BART-Encoder achieved competitive performance, with BERT obtaining the highest F1-score of 0.6342. Interestingly, decoder-based models such as LLaMA-3.1 and BART-Decoder also performed comparably well, suggesting that both architectures are viable for toxicity detection in English.
    
    \item For the Chinese \textbf{ToxiCN} dataset, we found notable differences in performance. While encoder models like Chinese BERT achieved reasonable scores (F1 = 0.5950), decoder-based models like Chinese LLaMA-3.1 and GLM-4 outperformed their encoder counterparts, with Chinese LLaMA-3.1 achieving the highest F1-score of 0.6165 and GLM-4 leading in silhouette score (0.4448). This demonstrates the language-specific effectiveness of decoder architectures.
    
    \item Decoder-based models tend to demonstrate superior clustering ability, as evidenced by higher Silhouette scores (e.g., GLM-4: 0.4448 and Chinese BART-Encoder: 0.5911), suggesting that their latent representations may be better suited for distinguishing toxic versus non-toxic content.
\end{itemize}

These findings reinforce the conclusion that both encoder and decoder models are effective for toxicity detection, though their relative effectiveness may vary depending on the dataset and language.

\subsection*{RQ2: How does expertise in toxicity understanding distribute across the layers of different language models?}

\begin{figure*}[t!]
    \centering
    \includegraphics[width=0.99\textwidth]{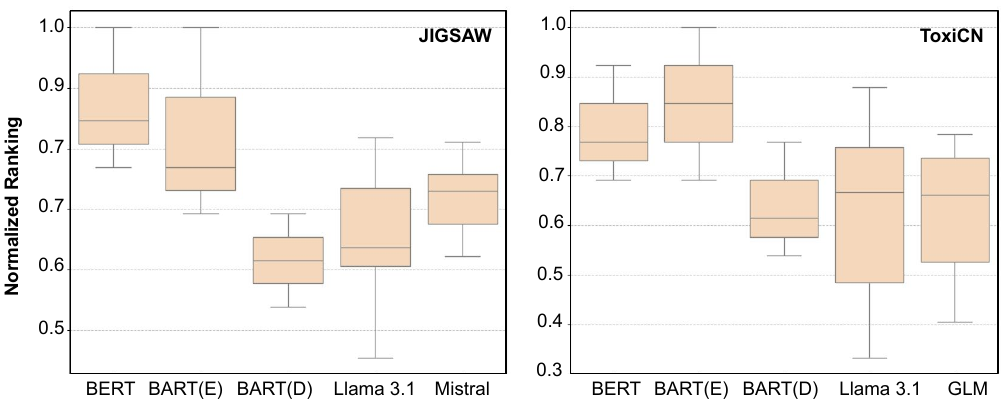}
        \caption{Layer-wise classification performance of different LMs, normalized for comparison with varying depths. The box plots illustrate the distribution of normalized rankings across layers, revealing where high-performing layers are concentrated. In encoder models: BERT and BART-Encoder, the top quartile (Q1) of high-performing layers is primarily located in the middle to end of the network, indicating that later layers are more specialized in toxic understanding. In contrast, decoder-based models: BART-Decoder, LLaMA-3.1, and GLM exhibit peak performance in the middle-to-late layers, as their final layers are optimized for text generation rather than classification.}
    \label{fig:box_plot}
    % \vspace{-5mm}
\end{figure*}

To explore this, we conducted a layer-wise expert analysis to identify where in the architecture the most effective toxic representations emerge. We normalized and ranked the classification performance across layers of different models, visualized in Figure~\ref{fig:box_plot}.

\begin{figure*}[t!]
    \centering
    \includegraphics[width=0.6\columnwidth]{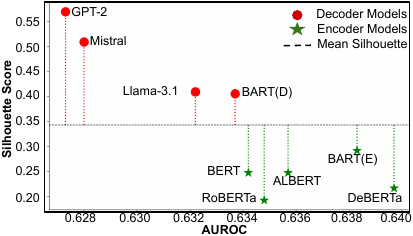}
        \caption{Relationship between AUROC and Silhouette Scores, highlighting a clear distinction between encoder and decoder models decoder models consistently achieve better clustering, challenging the common misconception that encoders are inherently superior for classification tasks.}
    \label{fig:auroc}
    % \vspace{-5mm}
\end{figure*}

Key findings include:

\begin{itemize}
    \item \textbf{Encoder models} (BERT, BART-Encoder): High-performing layers are predominantly located in the middle-to-late sections of the architecture. This suggests that toxicity-related semantic representations emerge and refine over the network's depth, with final layers contributing significantly to classification accuracy.
    
    \item \textbf{Decoder models} (BART-Decoder, LLaMA-3.1, GLM): These models exhibited peak layer-wise performance also in the mid-to-late stages, but notably not in the final layers. This aligns with their training objective—final layers are optimized for generation rather than classification, hence earlier layers are more informative for toxicity semantics.

    \item As depicted in Figure~\ref{fig:auroc}, there is a strong correlation between AUROC and Silhouette scores across models. Decoder models tend to achieve better clustering (higher silhouette scores), challenging the traditional belief that encoders are inherently better suited for classification tasks.
\end{itemize}

This structured analysis reveals that model expertise is not randomly distributed across layers. Instead, it tends to be localized in specific segments depending on the architecture, highlighting the value of understanding internal representation dynamics for improved model interpretability and downstream optimization.

\subsection*{Robustness to Catastrophic Forgetting}

A potential concern with suppressing generation experts is that such an intervention might induce catastrophic forgetting, thereby impairing the model’s performance on downstream tasks. To investigate this, we evaluated whether EigenShift preserves the general-purpose reasoning and factual knowledge abilities of the base models.

We conducted experiments on the Massive Multitask Language Understanding (MMLU) benchmark, which spans diverse domains including mathematics, logic, and factual knowledge. Table~\ref{tab:mmlu_results} reports the accuracy before and after intervention across representative MMLU subsets. The results demonstrate that while perplexity moderately increases, the semantic, reasoning, and knowledge-retention capabilities remain largely unaffected.

\begin{table}[h]
\centering
\begin{tabular}{lccc}
\toprule
\textbf{Domain} & \textbf{Base Model} & \textbf{Before} & \textbf{After} \\
\midrule
\multirow{5}{*}{Algebra} 
 & LLaMA   & 35 & 34 \\
 & Mistral & 29 & 32 \\
 & GPT-2   & 22 & 24 \\
 & Falcon  & 27 & 25 \\
 & MPT     & 22 & 21 \\
\midrule
\multirow{5}{*}{Logical Fallacies} 
 & LLaMA   & 46 & 46 \\
 & Mistral & 73 & 73 \\
 & GPT-2   & 19 & 18 \\
 & Falcon  & 31 & 30 \\
 & MPT     & 32 & 31 \\
\midrule
\multirow{5}{*}{U.S. Foreign Policy} 
 & LLaMA   & 59 & 58 \\
 & Mistral & 82 & 81 \\
 & GPT-2   & 28 & 29 \\
 & Falcon  & 32 & 30 \\
 & MPT     & 31 & 29 \\
\bottomrule
\end{tabular}
\caption{Accuracy on selected MMLU subsets before and after intervention. EigenShift preserves reasoning and factual knowledge with minimal degradation.}
\label{tab:mmlu_results}
\end{table}

These findings confirm that EigenShift’s intervention avoids catastrophic forgetting: the models maintain their reasoning, factual knowledge, and problem-solving skills despite the targeted suppression of generation experts. Thus, the trade-off of a moderate perplexity increase is both controlled and justifiable, contrasting with the severe degradations reported in prior work.

\subsection*{Scalability to Larger Language Models}

An important consideration for real-world applicability is whether EigenShift scales effectively to larger and more powerful language models. While smaller- and mid-scale models are useful for controlled analysis, deployment scenarios increasingly rely on high-capacity LLMs. To assess scalability, we extended our evaluation to include models ranging from 13B to 70B parameters.

Table~\ref{tab:scalability_results} reports the results on toxicity reduction and perplexity change. Across all tested models, EigenShift achieves substantial reductions in toxic generation, while maintaining perplexity within a moderate range. These results confirm that the method generalizes to larger-scale architectures without catastrophic degradation, further underscoring its practical relevance.

\begin{table}[h]
\centering
\resizebox{\linewidth}{!}{%
\begin{tabular}{lccccc}
\toprule
\textbf{Model} & \textbf{Tox. Before} & \textbf{Tox. After} & \textbf{PPL Before} & \textbf{PPL After} & \textbf{Tox. Red. / PPL $\Delta$} \\
\midrule
LLaMA-70B       & 11.20\% & 5.50\% & 4.49 & 4.70 & 50.88\% / -4.68\% \\
Falcon-30B      & 11.10\% & 4.39\% & 7.29 & 8.37 & 60.43\% / -14.81\% \\
LLaMA-13B       & 9.99\%  & 3.89\% & 5.69 & 7.83 & 61.06\% / -37.61\% \\
Mixtral 8$\times$7B & 10.85\% & 4.38\% & 4.93 & 4.99 & 59.64\% / -1.22\% \\
\bottomrule
\end{tabular}}
\caption{Scalability of EigenShift to larger-scale LLMs. The method substantially reduces toxicity while maintaining perplexity within acceptable ranges.}
\label{tab:scalability_results}
\end{table}